\documentclass{article}

\usepackage[nonatbib,final]{neurips_data_2023}

\usepackage[utf8]{inputenc} % allow utf-8 input
\usepackage[T1]{fontenc}    % use 8-bit T1 fonts
\usepackage{hyperref}       % hyperlinks
\usepackage[12pt]{moresize}
\usepackage{times}
\usepackage{environ} 
\usepackage{graphicx}
\usepackage{amsmath}
\usepackage{amssymb}
\usepackage{url} 
\usepackage{booktabs}
\usepackage{amsfonts}       % blackboard math symbols
\usepackage{nicefrac}       % compact symbols for 1/2, etc.
\usepackage{microtype}      % microtypography
\usepackage{array}
\usepackage{pifont}
\usepackage{capt-of,etoolbox}
\usepackage{multirow}
\usepackage{hhline}
\usepackage{bbding}
\usepackage{algorithm}
\usepackage{algorithmic}
\usepackage{boldline}
\usepackage{braket}
\usepackage{enumitem}
\usepackage{footnote}
\usepackage{wrapfig,lipsum}
\usepackage{makecell}
\usepackage[table,xcdraw]{xcolor}
\NewEnviron{NORMAL}{% 
    \scalebox{0.88}{$\BODY$} 
} 
\newcommand{\RNum}[1]{\uppercase\expandafter{\romannumeral #1\relax}}
\title{InSpaceType: Reconsider Space Type in Indoor Monocular Depth Estimation}
\author{%
  Cho-Ying Wu$^{1}$\quad Quankai Gao$^{1}$\quad Chin-Cheng Hsu$^{1}$ \\ \textbf{Te-Lin Wu}$^{2}$ \quad \textbf{Jing-Wen Chen}$^{1}$\quad \textbf{Ulrich Neumann}$^{1}$\\
  $^{1}$University of Southern California, $^{2}$University of California, Los Angeles\\
  \texttt{\{choyingw, quankaig, chincheh,jchen885,uneumann\}@usc.edu} \\ \texttt{\{telinwu\}@g.ucla.edu} \\
}

\begin{document}

\maketitle

Indoor monocular depth estimation has attracted increasing research interest. Most previous works have been focusing on methodology, primarily experimenting with NYU-Depth-V2 (NYUv2) Dataset, and only concentrated on the overall performance over the test set. However, little is known regarding robustness and generalization when it comes to applying monocular depth estimation methods to real-world scenarios where highly varying and diverse functional \textit{space types} are present such as library or kitchen. A study for performance breakdown into space types is essential to realize a pretrained model's performance variance. To facilitate our investigation for robustness and address limitations of previous works, we collect InSpaceType, a high-quality and high-resolution RGBD dataset for general indoor environments. We benchmark 12 recent methods on InSpaceType and find they severely suffer from performance imbalance concerning space types, which reveals their underlying bias. We extend our analysis to 4 other datasets, 3 mitigation approaches, and the ability to generalize to unseen space types. Our work marks the first in-depth investigation of performance imbalance across space types for indoor monocular depth estimation, drawing attention to potential safety concerns for model deployment without considering space types, and further shedding light on potential ways to improve robustness. See \url{https://depthcomputation.github.io/DepthPublic} for data and the supplementary document. The benchmark list on the GitHub project page keeps updates for the lastest monocular depth estimation methods.

%% older version 
% Indoor monocular depth estimation has attracted increasing research interest in recent years. 
% Previous works have been focusing on methodologies, 
% experimenting primarily on the NYU-Depth-V2 (NYUv2) dataset,
% and evaluated based on the overall performance over the entire test set. 
% However, little is known regarding robustness and generalization when it comes to applying depth estimation methods to real-world scenarios where a multitude of \textit{space types} are present.
% To address the limitations of previous works and to facilatate our investigation,
% we collected \textit{InSpaceType}, a high-quality 1242$\times$2208 resolution RGBD dataset for general indoor environments.
% We benchmarked 10 methods on InSpaceType and found them all suffer from performance imbalance with respect to space types.
% We extended our analysis to 4 other datasets, 3 mitigation approaches, and the ability to generalize to unseen space types.
% Our work marks the first in-depth investigation about performance imbalance across space types for indoor monocular depth estimation, shedding light on potential ways to improve robustness.

\section{Introduction}
\label{sec:intro}
Given an image input $I$, monocular depth estimation's target is to predict pixel-level depth map $D$ corresponding to $I$. It is a fundamental task in 3D vision for indoor applications, such as AR/VR gaming systems \cite{mehringer2022stereopsis,wu2021synergy,wu2016occlusion,wu2022cross,wu2018occluded,wu2021scene,wu2019salient}, robot assistance and navigation \cite{dong2022towards}, 3D photo creation \cite{shih20203d}, and novel view synthesis \cite{deng2022depth}. Challenges for indoor scenes lie in addressing highly diverse environments and arbitrarily arranged objects cluttered in the near field. Especially performances optimized for one environment may not apply to another due to highly varying structures between different space types of indoor environments. 

Most monocular depth estimation works begin from algorithmic perspectives, including advances in network architecture ~\cite{li2022depthformer,li2021structdepth,bian2021auto,jiang2021plnet,yuan2022new,kim2022global,bhat2022localbins,li2022binsformer,ramamonjisoa2021single,li2021structdepth,wu2020geometry}, loss function ~\cite{bhat2021adabins,fu2018deep,yin2019enforcing,liuva}, and learning paradigm on self-supervised learning ~\cite{wu2022toward, ji2021monoindoor, zhou2019moving, yu2020p}. NYU-Depth-V2 (NYUv2) pioneers in collecting indoor dense depth and thereafter becomes useful for monocular depth estimation in indoor scenes.

While being prominent in many prior studies, using NYUv2 as the only primary indoor depth estimation benchmark can bring potential shortcomings: (1) They always report error or accuracy numbers on the whole test set and overlook the variance of performance across different indoor \textit{space types}, which becomes a main concern in robustness when an edge-user applies pretrained models to uncommon or tailed space types and may observe degradation. 
Objects and textures are highly diverse for indoor scenes and may be specific to some spaces. For example, kitchenware is specific to kitchen and rarely appear in other spaces, or desk and chair are specific to the classroom.
Therefore, when a training set misses some space types, performances easily drop due to unseen objects or arrangements specific to those types. 
This issue becomes more serious because private spaces appear more often than other spaces in NYUv2. 
Without a breakdown across space types, the \textbf{practicability} of prior methods is still \textit{without verification}. (2) NYUv2 suffers from relatively low resolution (480$\times$640), an older camera imaging model, and high noise levels. These downsides make evaluations on NYUv2 less reliable and meet the needs of real high-quality applications.  

\begin{figure*}[bt!]
    \centering
    \includegraphics[width=0.95\linewidth]{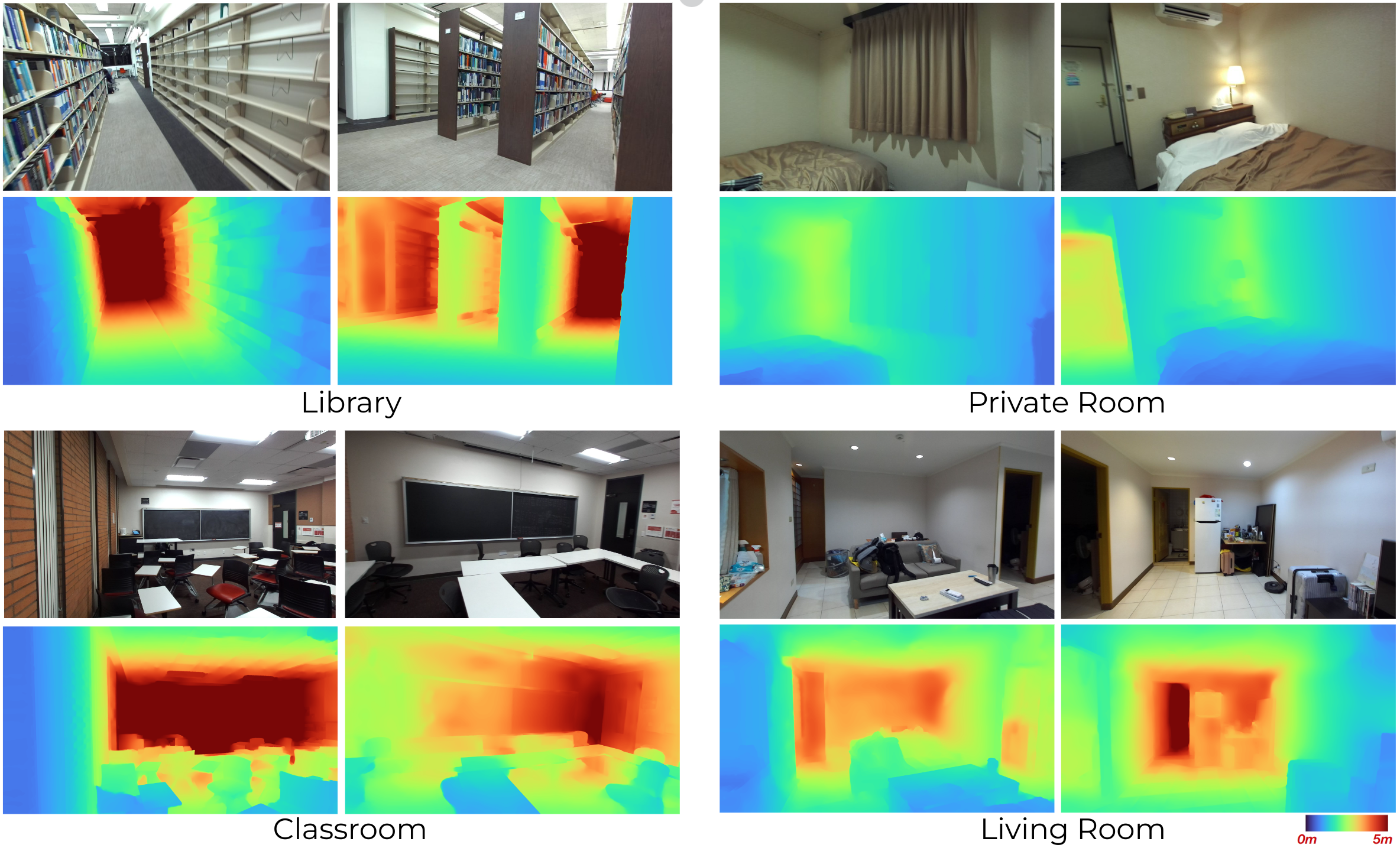}
    \vspace{-5pt}
    \caption{\textbf{Data samples of our InSpaceType Dataset.} }
    \vspace{-18pt}
    \label{fig_sample}
\end{figure*}

To enhance robustness and address the aforementioned problems, we take a deeper analysis of indoor space types using our novel dataset, \textit{InSpaceType, for benchmark and evaluation}. InSpaceType is collected by an off-the-shelf modern stereo camera system \cite{ZED2i} with a high resolution (1242$\times$2208), much less noise, high-quality depth maps aligned with images, and optimized performance to near-range depth sensing that is suitable for indoor scenes. We compare with other commonly used evaluation protocols for indoor monocular depth estimation in Table~\ref{table:dataset_comp}. 
% \footnote{Indoor datasets such as SUNRGBD~\cite{song2015sun}, ScanNet~\cite{dai2017scannet}, Hypersim~\cite{roberts2021hypersim}, and ARKitScenes~\cite{baruch1arkitscenes} are primarily for objection detection and segmentation, and Matterport3D~\cite{chang2017matterport3d} and Replica~\cite{straub2019replica} include 3D meshes and are mainly used in robotics applications such as navigation not prevalently used to evaluate monocular depth estimation performance}
Fig.~\ref{fig_sample} show data examples.

InSpaceType captures common space types, including private room, office, hallway, lounge, meeting room, large room, classroom, library, kitchen, playroom, living room, and bathroom. A hierarchical system is designed to describe these indoor space types. As the first step, 12 high-performing methods pretrained on NYUv2 are collected for InSpaceType zero-shot benchmarks, including supervised and self-supervised learning methods. The overall performances and type breakdown are exhibited. We find that those prior methods suffer from severe performance imbalance between space types. They perform well in head types such as private room ($\delta_1=92.05$) but much worse in tailed types such as large room ($\delta_1=54.93$). The presented type breakdown is practical as it \textit{goes beyond an average score and reveals performance variances on types}.
Our analysis helps us understand strength and weakness of a pretrained model and potentially reveals its underlying biases.

In addition to NYUv2, 3 other training datasets, including SimSIN (aggregation of Replica \cite{straub2019replica}, Matterport3D \cite{chang2017matterport3d}, and Habitat-Matterport3D), UniSIN \cite{wu2022toward}, and Hypersim \cite{roberts2021hypersim}, are experimented. We dig into characteristics in performance trained on these datasets and enumerate certain space types these datasets tend to have bias towards or against. In particular, we find synthetic or simulation datasets cannot accurately capture the intricate complexities of cluttered or small objects in real scenes, which is common in spaces such as kitchen.
To mitigate performance imbalance across types, 3 popular strategies are investigated: class re-weighting \cite{cui2019class,huang2016learning}, class-balanced sampling\cite{kangdecoupling}, and meta-learning\cite{wu2023meta,wu2019nonconvex}. 
Dividing the studied 12 types into 3 groups based on their spatial functions and then examining the generalizability between groups, we find generalization to unseen types is challenging. The best  $\delta_1$ accuracy can be as high as 98.12 for intra-group evaluation but drops to 59.02 in the worst case of inter-group evaluation.
%and examine generalizability between groups
%Last, we divide the studied twelve types into three groups based on their spatial functions and examine generalizability between groups. 
Our findings reveal that generalization to unseen groups is challenging due to high diversity of objects and mismatched scales across types.
Overall, this work serves a practical purpose for robustness and emphasizes the importance of the usually overlooked factor- space type in indoor environments.
We draw attention to potential safety concerns for model deployment without considering performance variance across space types.

Our contributions are summarized as follows:

\begin{itemize}[leftmargin=*,topsep=-3pt,itemsep=-0.0ex]
  \item To our best knowledge, we are the first to present a thorough analysis that considers space type in indoor monocular depth estimation. We benchmark 12 recent methods and reveal that they are biased towards/against certain types. We emphasize the importance of being aware of such bias for real-world applications. 
  \item We collect a dataset, InSpaceType, to facilitate our purpose of benchmarking and analyzing the variance of performances across multiple space types.
  \item We analyze 4 commonly-used training sets in indoor monocular depth estimation and enumerate their strengths and weaknesses towards certain space types. We further investigate 3 popular methods to mitigate performance imbalance.
\end{itemize}
\vspace{4pt}

%$\bullet$ We are the first to consider space type in indoor monocular depth estimation. We collect 10 high-performing methods for benchmarks and reveal that they suffer from underlying bias towards some space types.

%This echo previous findings regarding bias present in a pretrained model. One needs to attend to underlying bias and avoid apply to tailed types.

%various training datasets. It is crucial to attend to underlying bias present in a training dataset and avoid directly applying a pretrained model to its tailed type.

%This reinforces previous findings regarding the strengths and weaknesses of various training datasets. Therefore, it is crucial to pay attention to underlying biases present in the training dataset and avoid directly applying a pretrained model to a specific type without careful consideration.

\begin{table}[tb!]
\label{table:intra}
\begin{center}
  \caption{\textbf{Comparison of popular evaluation protocols for indoor monocular depth estimation.}}
  \vspace{-4pt}
  %\scriptsize
  %\footnotesize
  \ssmall
  \label{table:dataset_comp}
  \begin{tabular}[c]
  {
  p{1.4cm}<{\centering\arraybackslash}
  p{1.8cm}<{\centering\arraybackslash}
  p{0.9cm}<{\centering\arraybackslash}
  p{1.05cm}<{\centering\arraybackslash}
  p{1.0cm}<{\centering\arraybackslash}
  p{1.5cm}<{\centering\arraybackslash}
  p{4.00cm}<{\centering\arraybackslash}}
  \hlineB{2}
  
      Dataset  & Purpose & Resolution & Sensor &  Real or Synthetic & RGB-Imaging Quality &  Scene Diversity (\# of scenes, \# of RGBD pairs) \\
    \hline
      Diode ~\cite{vasiljevic2019diode} & Indoor + Outdoor & 1024$\times$768 & Lidar & Real & High & Very Low {\tiny (2 scenes, 753 indoor pairs)}\\
      IBims-1 ~\cite{koch2018evaluation} & Indoor focused & 640$\times$480 & Lidar & Real & Good & Low {\tiny (20 scenes, 100 pairs)}\\
      NYUv2 ~\cite{silberman2012indoor} &  Indoor focused & 640$\times$480 & Kinect-v1 & Real & Noisy & Medium {\tiny (private room focused, 654 pairs)}\\
      VA ~\cite{wu2022toward}  &  Indoor focused & 640$\times$640 & - & Synthetic & High & Very Low {\tiny (1 scene, 3523 pairs)} \\
      InSpaceType & Indoor focused & 2208$\times$1242 & Stereo & Real & High & High {\tiny (88 scenes, 1260 pairs)}\\
      
    \hlineB{2}
    \hline
  \end{tabular}
  \vspace{-15pt}
\end{center}
\end{table}
\section{Related Work}
\label{related}

\subsection{Indoor Monocular Depth Estimation} 
Monocular depth estimation is a fundamental task in computer vision.
Very early methods find cues in similar regions, shades, or motion to assign depth values \cite{dimiccoli2009monocular}, or use probabilistic models for depth estimation \cite{saxena2005learning}. This task has especially been popular during deep learning era. NYUv2 pioneers to collect a dense depth dataset for indoor scenes toward to goal. 

\textbf{Supervised method.}
Most methods operate in supervised learning and directly learn from paired RGBD data in the training set. 
Earlier deep methods include advances in architecture like fully convolutional neural network \cite{laina2016deeper} or operating in different learning paradigms such as multi-task learning \cite{eigen2015predicting}, transfer learning \cite{alhashim2018high}, dual-stream network \cite{two2017, zhong2019deep}, or multi-scale network \cite{moukari2018deep}. We organize more recent research directions as follows.

\vspace{-4pt}
$\bullet$ Loss design and discretized depth intervals: DORN \cite{fu2018deep} uses a space-increasing discretization strategy and recasts the depth regression as ordinal regression. Adabins \cite{bhat2021adabins} designs adaptive discretization and combines pixel-level regression loss and bin-center density loss. LocalBins \cite{bhat2022localbins} learns per-pixel discretization instead of global distribution.

\vspace{-4pt}
$\bullet$ Planarity: BTS \cite{lee2019big} use multi-scale planar guidance to estimate depth. P3Depth \cite{P3Depth} estimates plane coefficients for pieces and uses them for adaptive fusion.

\vspace{-4pt}
$\bullet$ Conditional random fields (CRFs): Early probabilistic approaches build on CRFs \cite{liu2015deep,ricci2018monocular}. Recently NeWCRFs \cite{yuan2022new} applies CRFs in windows to reduce computation overhead in fully connected CRFs.

\vspace{-4pt}
$\bullet$ Normal: Normal is helpful in depth regularization. VNL \cite{yin2019enforcing} enforces virtual normal constraints for depth prediction. IronDepth \cite{bae2022irondepth} uses normal map to propagate depth between pixels. 

\vspace{-4pt}
$\bullet$ Mixed-dataset training: MiDaS \cite{Ranftl2020} pioneers to collect 12 different mixed data source and results in high generalization of depth estimation. DPT \cite{Ranftl2021} further improves results using vision transformers. ZoeDepth \cite{bhat2023zoedepth} is based on DPT to further predict metric depth. LeReS \cite{yin2021learning,patakin2022single} also trains on mixed dataset for highly robust depth estimation. Depth-Anything~\cite{yang2024depth} proposes to use semantics and better training strategies to attain very high robustness.

\vspace{-4pt}
$\bullet$ Transformer: DepthFormer \cite{li2022depthformer} and PixelFormer \cite{agarwal2023attention} both use vision transformers and large models for higher accuracy. GLPDepth \cite{kim2022global} extracts global and local features with transformers and combines them with attention. Combined with transformers, MIM \cite{xie2023revealing} studies masked image modeling and builds upon large transformer models \cite{liu2022swin} for visual pretraining to learn good representation first and then finetune on downstream tasks. AiP-T \cite{ning2023all} uses VQGAN \cite{esser2021taming} and represents depth in a unified token space to attain high accuracy.  

\textbf{Self-Supervised method.} Most methods use NYUv2 and learn from consecutive frames with photometric consistency \cite{bian2021unsupervised, zhao2020towards, zhou2019moving}. DistDepth \cite{wu2022toward} adopts a distillation loss to learn from relative-depth pretrained models and incorporate left-right stereo consistency to learn metric depth.
%The advantage of self-supervised learning is it does not require depth groundtruth to train networks.

\subsection{Evaluation on Indoor Monocular Depth Estimation} 
Diode \cite{vasiljevic2019diode} collects both outdoor and indoor scenes. It collects high-quality data, but it contains very low diversity with only 2 scenes for evaluation. 
IBims-1 \cite{koch2018evaluation} is also limited in the amount of scenes.
VA \cite{wu2022toward} renders complex and high-quality images and depth maps but is only limited to one scene.
See Table ~\ref{table:dataset_comp} for the organization.

NYUv2 ~\cite{silberman2012indoor} is still popular in the evaluation of indoor monocular depth estimation. However, it is collected by an older Kinect-v1 system \cite{zhang2012microsoft} with noisy depth measurements and noisy imaging for RGB patterns. Its resolution is only 640$\times$480, which is much smaller than the needs of recent applications for robotics or high-resolution synthesis. Besides, NYUv2 mainly focuses on smaller and private rooms to evaluate performance. 
%Further, their camera viewing direction usually parallels the horizons, which limits the evaluation on robustness of viewing angles. 
%Our InspaceType collects data in more diverse viewing directions.
To overcome these limitations, our InSpaceType adopts a recent high-quality integrated stereo camera \cite{ZED2i} to collect high-resolution images and depth. 
%Outputs from the system are already rectified and well-aligned for images and depth maps. 
InSpaceType covers general-purpose and highly diverse indoor scenes, including private household spaces, workspaces, and campus scenes. We design a hierarchical system to describe space types and benchmark recent high-performing methods with detailed performance breakdown. This breakdown helps gain better understanding of performance variance across different spaces. 
%Further, one can realize speciality of the space types and further curate their training data for robust indoor estimation.

%Some methods operate in self-supervised studies. P$^2$Net \cite{yu2020p} estimates multiview consistency for plane in different time steps.  
\section{InSpaceType Dataset}
\label{dataset}
%The section introduces InSpaceType Dataset. We discuss in detail the data collection process, statistics, and present train and test splits.

We capture images and depth maps computed from left-right stereo pairs using a high-quality integrated stereo camera system, Zed-2i \cite{ZED2i}.
Its baseline is 12cm, field of view (FOV) is $120^\circ$, working distance is up to 20m, and its backend engine enables dense depth outputs

It is particularly optimized for ranges within 15m, whose average error is within 5$\%$ from its specification. This matches our needs to work in indoor environments. 
We operate in its ULTRA mode to output a high resolution of 2208$\times$1242. 
The stereo camera device is anchored on a hand-held stabilizer during data collection.
Wide FOV makes scene images contain more cues to estimate distance. We do not zoom into small objects or flat walls, which will cause ambiguity in the scene scale. The pitch angle is about within $\pm 30^\circ$, and the roll angle is within $\pm 10^\circ$. This setting enriches scenes captured from different viewing directions without resulting in strange scenes or reaching a threshold where excessively large angles would eliminate cues that indicate depth ranges.
For better quality, we also avoid non-Lambertian areas such as mirrors or highly reflective surfaces.
%To avoid non-Lambertian areas, which will break down depth measurement, we avoid mirrors or highly reflective areas.

Our environments cover household spaces, workspaces, and campus spaces, including private room, office, hallway, lounge, meeting room, large room, classroom, library, kitchen. 88 different environments are visited in total. We record at 15fps while walking around those spaces. Around 40K images are collected. To create the evaluation set, we manually select 1260 images from all the environments.

Our selection criteria include (1) clear imaging with minimal motion blur, (2) not selecting from nearby 10 frames, and (3) containing sufficient cues that hint depth scales of the scene.
Fig.~\ref{fig_stat} shows the dataset statistics. See Supplementary for more dataset descriptions.

\begin{figure*}[bt!]
    \centering
    \includegraphics[width=1.0\linewidth]{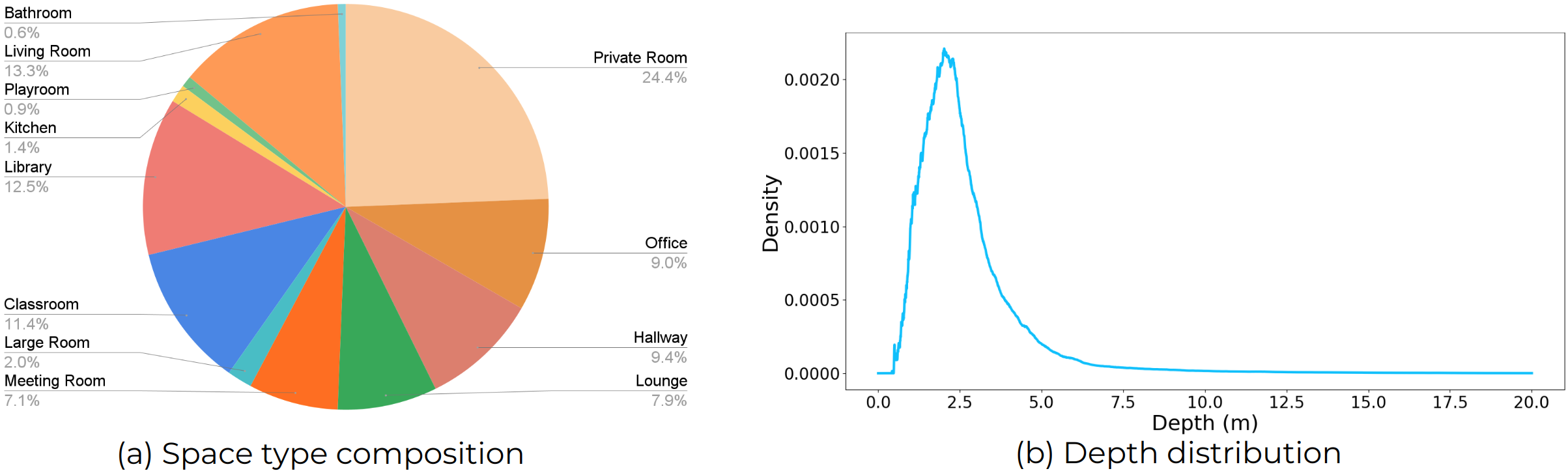}
    \vspace{-12pt}
    \caption{\textbf{Statistics for InSpaceType evaluation set.} }
    \vspace{-4pt}
    \label{fig_stat}
\end{figure*}

\section{Cross-Dataset Benchmarks}
\label{benchmark}

%In the following section, we show several analysis and discussion based on InSpaceType.

\textbf{[\RNum{1}: Benchmarks]} As the first step, we collect recent 11 high-performing methods and fetch open-sourced models pretrained on NYUv2 and use them to make inferences on InSpaceType. The following methods are included: DPT \cite{Ranftl2021}, GLPDepth \cite{kim2022global}, AdaBins \cite{bhat2021adabins}, PixelFormer \cite{agarwal2023attention}, NeWCRFs \cite{yuan2022new}, BTS \cite{lee2019big}, MIM \cite{xie2022revealing}, IronDepth \cite{bae2022irondepth}, Decomposition \cite{jun2022depth}, ZoeDepth \cite{bhat2023zoedepth}, Depth-Anything~\cite{yang2024depth}\footnote[1]{We use NYUv2 fine-tuned indoor model for evaluation. See \url{https://github.com/DepthComputation/InSpaceType_Benchmark} for full space type breakdown}. We adopt error (AbsRel, SqRel, RMSE) and accuracy metrics ($\delta_1$, $\delta_2$, $\delta_3$ with base factor 1.25) commonly used in the literature of monocular depth estimation for evaluation. To compensate for different camera intrinsics between NYUv2 and InSpaceType, we follow prior protocols for cross-dataset evaluation \cite{luo2020consistent, wu2023meta} and use median-scaling to calibrate prediction and groundtruth scale.

\begin{table}[tb!]
\centering
\caption{\textbf{InSpaceType benchmark: overall performance.} The best number is in bold, and the second-best is underlined. We include ten recent high-performing methods, including supervised and self-supervising learning paradigms.}
\vspace{-6pt}
\ssmall
\label{table:2}
\begin{tabular}{|c|c|c|c|c|c|c|c|c|c|c|}
\hline
\cellcolor[HTML]{99CCFF} Method & \cellcolor[HTML]{99CCFF} Year &\cellcolor[HTML]{99CCFF} Architecture & \cellcolor[HTML]{FAE5D3} MAE & \cellcolor[HTML]{FAE5D3} AbsRel & \cellcolor[HTML]{FAE5D3} SqRel & \cellcolor[HTML]{FAE5D3} RMSE & \cellcolor[HTML]{D5F5E3} $\delta_1$ & \cellcolor[HTML]{D5F5E3} $\delta_2$ & \cellcolor[HTML]{D5F5E3} $\delta_3$ \\
\hline
\multicolumn{10}{c}{\cellcolor[HTML]{FFFE65}Supervised Learning}\\
BTS \cite{lee2019big}& arXiv'19 & DenseNet-161 & 0.3602 & 0.1445 & 0.1162 & 0.5222 & 81.65 & 95.57 & 98.54 \\
AdaBins \cite{bhat2021adabins}& CVPR'21 & Unet+AdaBins & 0.3341 & 0.1333 & 0.0957 & 0.4922 & 83.64 & 96.36 & 98.92 \\
DPT \cite{Ranftl2021} & ICCV'21 & DPT-Hybrid & 0.3090 & \underline{0.1224} & 0.0773 & 0.4616 & 85.96 & 97.17 & \underline{99.19} \\
GLPDepth \cite{kim2022global}& arXiv'22 & Mit-b4 & 0.3068 & 0.1239 & 0.0788 & 0.4527 & 86.05 & \underline{97.36} & 99.16 \\
IronDepth \cite{bae2022irondepth}& BMVC'22 & EfficientNet-B5 & 0.3271 & 0.1276 & 0.1022 & 0.4894 & 85.30 & 96.37 & 98.84 \\
Decomposition \cite{jun2022depth} & ECCV'22 & EfficientNet-B5 & 0.3274 & 0.1278 & 0.1025 & 0.4899 & 85.25 & 96.35 & 98.83 \\
NeWCRFs \cite{yuan2022new}& CVPR'22 & Swin-Large & 0.3028 & 0.1251 & 0.0823 & 0.4541 & 86.04 & 96.68 & 98.94 \\
PixelFormer \cite{agarwal2023attention} & WACV'23 & Swin-Large & \underline{0.2982} & 0.1225 & \underline{0.0761} & \underline{0.4392} & \underline{86.08} & 97.03 & 99.10 \\
MIM \cite{xie2022revealing}& CVPR'23 & SwinV2-Large & 0.2807 & 0.1100 & 0.0679 & 0.4244 & 88.58 & 97.59 & 99.28 \\
ZoeDepth (NK) \cite{bhat2023zoedepth}& arXiv'23 & BeiT-Large & 0.2469 & 0.0969 & 0.0527 & 0.3834 & 90.76 & 98.19 & 99.50 \\
ZoeDepth (N) \cite{bhat2023zoedepth}& arXiv'23 & BeiT-Large & 0.2484 & 0.0956 & 0.0528 & 0.3887 & \textbf{90.81} & \textbf{98.22} & \textbf{99.52} \\
DepthAnything \cite{yang2024depth}& arXiv'24 & ViT-L & \textbf{0.2397} & \textbf{0.0928} & \textbf{0.0506} & \textbf{0.3806} & 90.01 & 98.09 & \textbf{99.54} \\
\hline
\multicolumn{10}{c}{\cellcolor[HTML]{FFFE65}Self-Supervised Learning}\\
DistDepth  $_\text{{(DPT-Hybrid)}}$ \cite{wu2022toward} & CVPR'22 & ResNet152  & 0.4688 & 0.1746 & 0.1718 & 0.6877 & 74.71 & 94.18 & 98.60 \\
DistDepth  $_\text{{(DPT-Large)}}$  \cite{wu2022toward} & CVPR'22 & ResNet152  & 0.3817 & 0.1447 & 0.1094 & 0.5758 & 81.05 & 95.46 & 98.69 \\
\hline
\end{tabular}
\vspace{-5pt}
\end{table}

\begin{figure*}[bt!]
    \centering
    \includegraphics[width=1.0\linewidth]{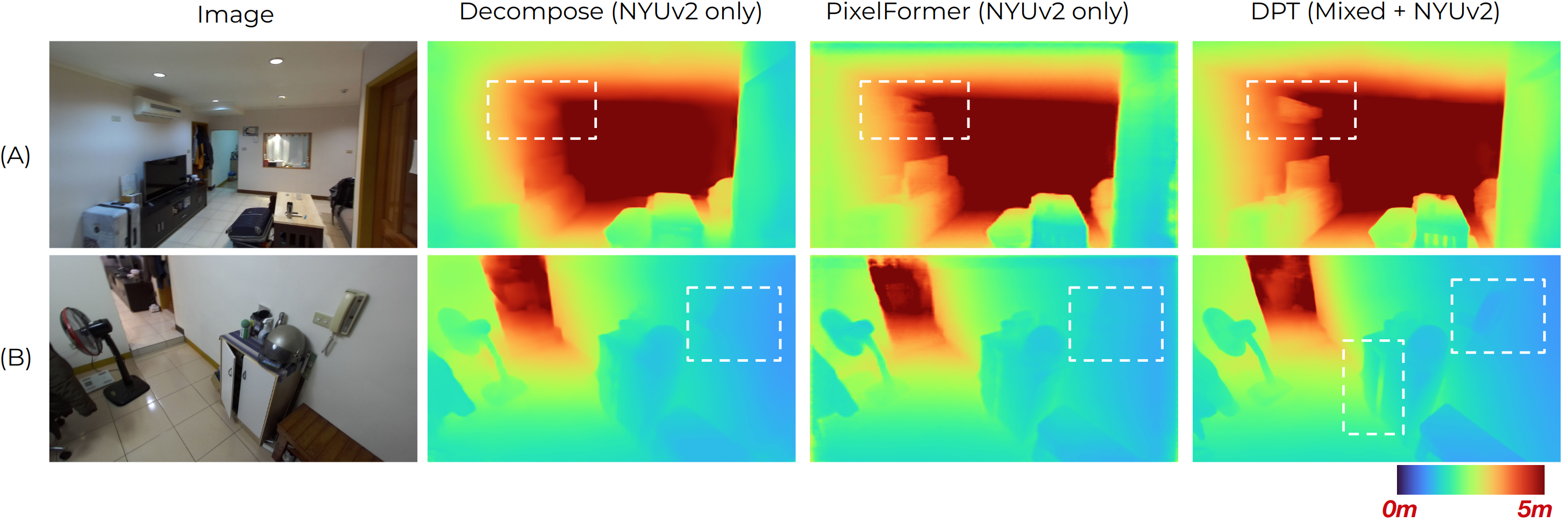}
    \vspace{-20pt}
    \caption{\textbf{Regions that trained on NYUv2 only cannot show}. InSpaceType contains several object arrangements that NYUv2 does not include, such as wall-hanging air-conditioner and phone are mostly exclusive to Asian styles rooms;   
    a tilted viewing direction for pitch angle is shown in (B), where training on NYUv2 only cannot give robust results because NYUv2 has minor viewing pitch angle changes. 
    DPT in their setting trains on 10 different dataset + NYUv2 (mixed-set training)  attains more pleasant results.
    To verify generalizability, InSpaceType serves as a testbed to help find out cases where training on popular NYUv2 only cannot show.}
    \vspace{-4pt}
    \label{fig_diff}
\end{figure*}

Table ~\ref{table:2} shows the results. For reference, the performance for these methods on NYUv2 benchmark (in terms of lower RMSE) is listed: Depth-Anything > ZoeDepth > MIM > PixelFormer > NeWCRFs > GLPDepth > IronDepth > Decomposition > DPT > AdaBins > BTS. The ranking on InSpaceType is consistent to NYUv2. Depth-Anything, ZoeDepth, MIM and PixelFormer are top among published methods. They use large-size transformers, showing large models can learn better representations. 
We also notice DPT surpasses other methods in some metrics. This is because DPT was first pretrained on a mixture of datasets and then finetuned on NYUv2. Larger amount of data involved during pretraining helps the model to learn better representations for generalization. To verify the generalizability in Fig.~\ref{fig_diff}, we show examples where training only on NYUv2 may stumble.

Self-supervised direction attracted more attention due to no depth groundtruth involved during training but generally lower performance, especially in driving scenarios.
Indoor self-supervised methods received less attention due to the lack of indoor stereo pairs to enable robust monocular depth scales. DistDepth \cite{wu2022toward} pioneers to create synthetic indoor stereo data and enable more stable scale estimation from monocular images and becomes current SOTA. 
We adopt DistDepth to show the current performance gap between supervised/ self-supervised learning. The gap tells us  
how far it is between training with/ without depth groundtruth.

%Comparing supervised and self-supervised learning, one can find supervised learning leads to much more robust generalization since supervised methods have access to groundtruth depth during training. 

\textbf{[\RNum{2}: Breakdown by Space Type]} Then, we break down MIM and PixelFormer's performances by space types and show results in Table ~\ref{table:25}. 
%This can potentially reveal the underlying distribution for space types in NYUv2's training dataset.
%\caption{\textbf{Performance breakdown by space types.} Both supervised-learning and self-supervised learning are investigated. We study MIM using SwinV2-Large pretrained on NYUv2 the former, and DistDepth$_\text{{(DPT-Large)}}$ using ResNet152 pretrained on SimSIN for the latter.
%Beside the breakdown, we also list top-5 types based on lower/higher error (RMSE) and accuracy ($\delta_1$).}
\begin{table}[bt!]
\centering
\caption{\textbf{Performance breakdown by space types.} We study MIM and PixelFormer, which are top-performing published methods.
Beside the breakdown, we also list top-5 space types based on lower/higher error (RMSE) and accuracy ($\delta_1$). Easy and hard type are listed based on co-occurrence.}
\vspace{-6pt}
\tiny
%\footnotesize
\label{table:25}
\begin{tabular}{|c|c|c|c|c|c|c|c|c|c|c|c|c|}
\hline
& \multicolumn{6}{c}{\cellcolor[HTML]{FFFE65}MIM Results} & \multicolumn{6}{c}{\cellcolor[HTML]{F7BCDC}PixelFormer Results}  \\

\cellcolor[HTML]{99CCFF} Type & \cellcolor[HTML]{FAE5D3} AbsRel & \cellcolor[HTML]{FAE5D3} SqRel & \cellcolor[HTML]{FAE5D3} RMSE & \cellcolor[HTML]{D5F5E3} $\delta_1$ & \cellcolor[HTML]{D5F5E3} $\delta_2$ & \cellcolor[HTML]{D5F5E3} $\delta_3$ & \cellcolor[HTML]{FAE5D3} AbsRel & \cellcolor[HTML]{FAE5D3} SqRel & \cellcolor[HTML]{FAE5D3} RMSE & \cellcolor[HTML]{D5F5E3} $\delta_1$ & \cellcolor[HTML]{D5F5E3} $\delta_2$ & \cellcolor[HTML]{D5F5E3} $\delta_3$ \\
\hline
Private room & 0.0927 & 0.0342 & 0.2556 & 92.05 & 98.99 & 99.83 & 0.1013 & 0.0372 & 0.2638 & 90.17 & 98.70 & 99.80 \\
Office  & 0.1106 & 0.0532 & 0.3313  & 87.67 & 97.42 & 99.44 & 0.1271 & 0.0649 & 0.3658 & 84.60 & 96.38 & 99.19 \\ 
Hallway & 0.1229 & 0.0805 & 0.5463  & 85.66 & 96.52 & 98.89 & 0.1418 & 0.0880 & 0.5236 & 81.46 & 96.40 & 99.05\\ 
Lounge & 0.1316 & 0.1290 & 0.7447 & 84.15 & 96.22 & 98.93 & 0.1500 & 0.1624 & 0.8215 & 79.92 & 94.79 & 98.41 \\
Meeting room & 0.0984 & 0.0483 & 0.3636 & 91.70 & 98.69 & 99.64 & 0.1103 & 0.0551 & 0.3873 & 88.25 & 98.57 & 99.76\\ 
Large room & 0.2683 & 0.4499 & 1.3915 & 54.93 & 83.89 & 93.91 & 0.2125 & 0.3119 & 1.1396 & 67.71 & 88.59 & 95.40 \\ 
Classroom & 0.0781 & 0.0334 & 0.3071 & 94.52 & 99.25 & 99.83 & 0.0912 & 0.0403 & 0.3368 & 91.37 & 99.14 & 99.86 \\
Library & 0.1342 & 0.0978 & 0.6281 & 85.61 & 96.57 & 98.57 & 0.1530 & 0.1242 & 0.6527 & 82.57 & 94.88 & 97.73\\
Kitchen & 0.1482 & 0.0791 & 0.3374 & 82.31 & 95.42 & 98.32 & 0.1819 & 0.0978 & 0.3521 & 78.86 & 93.05 & 96.63\\ 
Playroom & 0.0702 & 0.0276 & 0.2466 & 94.54 & 98.25 & 99.79 & 0.1010 & 0.0476 & 0.3042 & 92.11 & 97.23 & 99.15 \\ 
Living room & 0.1033 & 0.0502 & 0.3448 & 89.07 & 97.88 & 99.52 & 0.1132 & 0.0568 & 0.3601 & 87.35 & 97.31 & 99.34  \\
Bathroom & 0.1456 & 0.0772 & 0.2788 & 83.45 & 96.07 & 98.13 & 0.1439 & 0.0570 & 0.2488 & 83.72 & 96.55 & 98.36\\
\hline
\end{tabular}

\begin{tabular}{|l|l|l|}

\hline
& \multicolumn{1}{c}{\cellcolor[HTML]{FFFE65}MIM} & \multicolumn{1}{c}{\cellcolor[HTML]{F7BCDC} PixelFormer}  \\
\hline
\cellcolor[HTML]{FAE5D3} Top-5 Lower RMSE & playroom, private room, bathroom, classroom, office & bathroom, private room, playroom, classroom, kitchen\\
\cellcolor[HTML]{FAE5D3} Top-5 Higher $\delta_1$ & playroom, classroom, private room, meeting room, living room & playroom, private room, classroom, meeting room, living room \\
\hline
\multicolumn{1}{|c|} {\cellcolor[HTML]{FAE5D3} Easy type} & \cellcolor[HTML]{FAE5D3} playroom, private room, classroom & \cellcolor[HTML]{FAE5D3} playroom, private room, classroom \\
\hline
\cellcolor[HTML]{D5F5E3} Top-5 Higher RMSE & large room, lounge, library, hallway, meeting room & large room, lounge, library, hallway, meeting room\\
\cellcolor[HTML]{D5F5E3} Top-5 Lower $\delta_1$ & large room, kitchen, bathroom, lounge, library & large room, kitchen, lounge, hallway, library\\
\hline
\multicolumn{1}{|c|} {\cellcolor[HTML]{D5F5E3} Hard type} & \cellcolor[HTML]{D5F5E3} large room, lounge, library & \cellcolor[HTML]{D5F5E3} large room, lounge, library, hallway \\
\hline
\end{tabular}
\vspace{-9pt}
\end{table}
We show performance on each space type, list out top-5 high/ low-performing types, and show easy/ hard types based on the concordance in lower error/ higher accuracy (easy) and the opposite (hard). 
The easy types for MIM and PixelFormer are the same: playroom, private room, and classroom; the common hard types are: large room, lounge, and library. 
Clear easy and hard types indicate that strength and weakness for these models are apparent, showing they are biased towards/against some specific types.
The easy and hard types are highly overlapping for both MIM and PixelFormer, which further unveils potential bias underlying NYUv2's training data.
The most frequent space type for NYUv2 is private room and living room, which are typically small spaces. 
By contrast, large spaces of farther ranges are those less capable of.

%In Fig. ~\ref{zs_stat} we show dataset statistics of NYUv2 using our defined labels. 
%NYUv2 has living room and private room as the most frequent scenes, which are typical small rooms. 

%The easy and hard types match the distribution of NYUv2.
%The easy and hard types indicate where is the best to use the pretrained model or where one needs to avoid.
From the above analysis, we find performance varies a lot across different space types. The identification of easy and hard types provides insights into the suitability of using pretrained models in specific scenarios or the need to avoid them. Furthermore, this analysis highlights that the learned representations from NYUv2 still have a gap to transfer to other space types. See Supplementary for performance breakdown for other methods and hierarchical description for types.

\textbf{[\RNum{3}: More Training Dataset Generalization]} Next, we validate performances trained on other popular training datasets. Specifically, we include the following datasets and models: SimSIN \cite{wu2022toward}(self-supervised by DistDepth, ResNet152 \cite{he2016deep}), UniSIN \cite{wu2022toward} (self-supervised by DistDepth, ResNet152), Hypersim \cite{roberts2021hypersim} (supervised, ConvNeXt-Base \cite{liu2022convnet}). 
SimSIN and UniSIN datasets are recently introduced along with DistDepth \cite{wu2022toward} with pretrained models released on both. Both datasets mainly serve the purpose of self-supervised studies that recently become popular. Thus, we consider including such self-supervised oriented datasets, SimSIN and UniSIN, and use DistDepth pretrained model for analysis. 
Table~\ref{table:3} and ~\ref{table:4} show the results. We show depth distribution of these datasets in Supplementary for reference.

\begin{table}[bt!]
\centering

\caption{\textbf{Performance trained on SimSIN and UniSIN.} We leverage pretrained DistDepth$_\text{{(DPT-Large)}}$-ResNet152 models ~\cite{wu2022toward} in both cases.}
\vspace{-5pt}
%\ssmall
\tiny
\label{table:3}
\begin{tabular}{|c|c|c|c|c|c|c|c|c|c|c|c|c|}
\hline
& \multicolumn{6}{c}{\cellcolor[HTML]{FFFE65}Trained on SimSIN} & \multicolumn{6}{c}{\cellcolor[HTML]{F7BCDC}Trained on UniSIN}  \\

\cellcolor[HTML]{99CCFF} Type & \cellcolor[HTML]{FAE5D3} AbsRel & \cellcolor[HTML]{FAE5D3} SqRel & \cellcolor[HTML]{FAE5D3} RMSE & \cellcolor[HTML]{D5F5E3} $\delta_1$ & \cellcolor[HTML]{D5F5E3} $\delta_2$ & \cellcolor[HTML]{D5F5E3} $\delta_3$ & \cellcolor[HTML]{FAE5D3} AbsRel & \cellcolor[HTML]{FAE5D3} SqRel & \cellcolor[HTML]{FAE5D3} RMSE & \cellcolor[HTML]{D5F5E3} $\delta_1$ & \cellcolor[HTML]{D5F5E3} $\delta_2$ & \cellcolor[HTML]{D5F5E3} $\delta_3$ \\
\hline
Private room  & 0.1509 & 0.0986 & 0.4447 & 79.36 & 96.48 & 99.53 & 0.1738 & 0.1102 & 0.4539 & 72.81 & 93.65 & 98.68\\
Office & 0.1812 & 0.1606 & 0.5789 & 74.01 & 93.47 & 97.80 & 0.1656 & 0.1113 & 0.5020 & 75.04 & 94.86 & 98.98 \\ 
Hallway & 0.1597 & 0.1239 & 0.6324 & 78.12 & 94.92 & 98.64 & 0.1262 & 0.0765 & 0.4977 & 84.79 & 96.91 & 99.37 \\ 
Lounge & 0.1841 & 0.2148 & 0.9037 & 73.86 & 93.11 & 97.94 & 0.1380 & 0.1428 & 0.7351 & 82.41 & 96.77 & 98.95 \\ 
Meeting room & 0.1962 & 0.2491 & 0.9417  & 66.91 & 93.58 & 99.45 & 0.1300 & 0.1104 & 0.5937 & 83.94 & 97.36 & 99.63 \\ 
Large room  & 0.1842 & 0.2619 & 1.0727 & 72.79 & 91.87 & 97.88 & 0.1608 & 0.2112 & 0.9447 & 75.95 & 94.83 & 98.68 \\ 
Classroom  & 0.2069 & 0.3288 & 1.0292  & 67.11 & 91.66 & 98.51 & 0.1313 & 0.1166 & 0.6077 & 84.12 & 97.96 & 99.74 \\
Library & 0.1857 & 0.1913 & 0.8307 & 75.14 & 93.11 & 97.44 & 0.1230 & 0.1146 & 0.6958 & 85.61 & 96.48 & 98.91 \\ 
Kitchen  & 0.2524 & 0.2083 & 0.5649 & 59.22 & 87.44 & 96.12 & 0.2741 & 0.1997 & 0.5740 & 52.30 & 85.15 & 96.28 \\ 
Playroom & 0.1597 & 0.1147 & 0.5946 & 75.59 & 97.95 & 99.67 & 0.1486 & 0.0822 & 0.4755 & 78.63 & 98.23 & 99.84 \\ 
Living room & 0.1600 & 0.1166 & 0.5284 & 77.05 & 94.90 & 98.89 & 0.1644 & 0.1153 & 0.5106 & 76.37 & 93.66 & 98.48 \\
Bathroom & 0.1751 & 0.1153 & 0.3900 & 74.22 & 94.07 & 98.46 & 0.1384 & 0.0409 & 0.2168 & 84.67 & 95.28 & 99.68 \\
\hline
All & 0.1746 & 0.1719 & 0.6877 & 74.72 & 94.18 & 98.61 & 0.1509 & 0.1143 & 0.5602 & 78.96 & 95.38 & 98.97\\
\hline
\end{tabular}

\begin{tabular}{|l|l|l|}
\hline
& \multicolumn{1}{c}{\cellcolor[HTML]{FFFE65} SimSIN} & \multicolumn{1}{c}{\cellcolor[HTML]{F7BCDC} UniSIN}  \\
\hline
\cellcolor[HTML]{FAE5D3} Top-5 Lower RMSE & bathroom, private room, living room, kitchen, office & bathroom, private room, playroom, hallway, office\\
\cellcolor[HTML]{FAE5D3} Top-5 Higher $\delta_1$ & private room, hallway, living room, playroom, library & library, hallway, bathroom, classroom, meeting room \\
\hline
\multicolumn{1}{|c|} {\cellcolor[HTML]{FAE5D3} Easy type} & \cellcolor[HTML]{FAE5D3} living room, private room & \cellcolor[HTML]{FAE5D3} bathroom, hallway \\
\hline
\cellcolor[HTML]{D5F5E3} Top-5 Higher RMSE & large room, classroom, meeting room, lounge, library & large room, lounge, library, classroom, meeting room\\
\cellcolor[HTML]{D5F5E3} Top-5 Lower $\delta_1$ & kitchen, meeting room, classroom, large room, lounge & kitchen, private room, office, large room, living room\\
\hline
\multicolumn{1}{|c|} {\cellcolor[HTML]{D5F5E3} Hard type} & \cellcolor[HTML]{D5F5E3} large room, classroom, meeting room, lounge & \cellcolor[HTML]{D5F5E3} large room \\
\hline
\end{tabular}
\vspace{8pt}
\centering
\caption{\textbf{Performance trained on Hypersim.} We adopt supervised learning with standard $L_2$ loss on a ConvNeXt backbone \cite{liu2022convnet}. }
\vspace{-6pt}
\tiny
\label{table:4}
\begin{tabular}{|c|c|c|c|c|c|c|c|c|}
\hline
& \multicolumn{6}{c}{\cellcolor[HTML]{FFFE65}Hypersim}  \\
\hline
\cellcolor[HTML]{99CCFF} Type & \cellcolor[HTML]{FAE5D3} AbsRel & \cellcolor[HTML]{FAE5D3} SqRel & \cellcolor[HTML]{FAE5D3} RMSE & \cellcolor[HTML]{D5F5E3} $\delta_1$ & \cellcolor[HTML]{D5F5E3} $\delta_2$ & \cellcolor[HTML]{D5F5E3} $\delta_3$ \\
\hline

Private room  & 0.1321 & 0.0607 & 0.3376 & 84.55 & 97.26 & 99.36 \\
Office & 0.1678 & 0.1085 & 0.4916 & 76.01 & 93.60 & 97.87 \\
Hallway  & 0.2126 & 0.1912 & 0.8347 & 65.67 & 89.89 & 97.15 \\
Lounge  & 0.1937 & 0.2254 & 0.9403 & 71.31 & 92.58 & 97.58 \\
Meeting room  & 0.1315 & 0.0837 & 0.5333  & 81.49 & 98.23 & 99.74 \\
Large room  & 0.2525 & 0.4125 & 1.3897  & 56.48 & 85.13 & 94.95 \\
Classroom  & 0.1258 & 0.0740 & 0.4700  & 85.07 & 98.50 & 99.80 \\
Library  & 0.1766 & 0.1664 & 0.8654  & 73.53 & 93.54 & 98.03 \\
Kitchen  & 0.2417 & 0.1343 & 0.4347 & 62.81 & 90.25 & 96.43 \\
Playroom & 0.1629 & 0.1132 & 0.5663  & 77.09 & 93.87 & 99.15 \\
Living room  & 0.1485 & 0.0872 & 0.4519 & 80.91 & 95.40 & 98.95 \\
Bathroom  & 0.1648 & 0.0627 & 0.2908 & 76.11 & 95.80 & 98.89 \\
\hline
All  & 0.1592 & 0.1187 & 0.5803 & 77.95 & 94.99 & 98.60 \\
\hline
\end{tabular}
\vspace{5pt}
\begin{tabular}{|l|l|}

\hline
& \multicolumn{1}{c}{\cellcolor[HTML]{FFFE65} Hypersim}   \\
\hline
\cellcolor[HTML]{FAE5D3} Top-5 Lower RMSE & bathroom, private room, kitchen, living room, classroom \\
\cellcolor[HTML]{FAE5D3} Top-5 Higher $\delta_1$ & classroom,  private room, meeting room, living room, playroom \\
\hline
\multicolumn{1}{|c|} {\cellcolor[HTML]{FAE5D3} Easy type} & \cellcolor[HTML]{FAE5D3} private room, living room, classroom \\
\hline
\cellcolor[HTML]{D5F5E3} Top-5 Higher RMSE & large room, lounge, library, hallway, playroom \\
\cellcolor[HTML]{D5F5E3} Top-5 Lower $\delta_1$ & large room, kitchen, hallway, lounge, library\\
\hline
\multicolumn{1}{|c|} {\cellcolor[HTML]{D5F5E3} Hard type} & \cellcolor[HTML]{D5F5E3} large room, library, hallway, lounge  \\
\hline
\end{tabular}
\vspace{-18pt}
\end{table}

\vspace{-4pt}
$\bullet$ SimSIN: It contains data from Replica \cite{straub2019replica}, Matterport3D \cite{chang2017matterport3d}, and HM3D \cite{ramakrishnan2021habitat}, which are also focused on household spaces. From Table ~\ref{table:3}, its easy types are private room and living room, and hard types are large room, classroom, meeting room, and lounge. Its strength and weakness are also obvious, showing that SimSIN is heavily biased towards household spaces, and is especially under-performing in workspaces or campus scenes. 

\vspace{-4pt}
$\bullet$ UniSIN: From Table ~\ref{table:3}, its easy types are bathroom and hallway, and hard type is only large room. One can observe UniSIN has less bias towards space types, having only few easy and hard types. We assume it is because UniSIN collects data from more diverse  environments and avoids clear bias.

\vspace{-4pt}
$\bullet$ Hypersim: From Table ~\ref{table:4}, its easy types are private room, classroom, living room, and its hard types are large room, library, hallway, and lounge. It also has obvious bias, especially bias towards household spaces and classroom and bias against large room or types missed in the dataset such as library or hallway. Though Hypersim contains high-quality renderings from synthetic environments, it also focuses on household spaces and biases against several common space types.

%The above descriptions serve as a datasetsheet to assist in understanding what each dataset focuses on. 
One can find SimSIN and Hypersim, both are rendered from simulation platforms, have more obvious bias types, compared with UnSIN of real-world data collection. This indicates current trends to curate indoor synthetic data focus more on head types, especially private room and living room as the most frequent application scenarios, and may miss tailed types such as library, lounge, or hallway that is common but easily overlooked. Deploying models trained on those datasets may not be robustness in the wild. We point out this observation to call for attention when curating synthetic datasets.

\vspace{-4pt}
$\bullet$ Special Type: We find kitchen is a special type, which is of lower RMSE but also very low accuracy score $\delta_1$ in SimSIN and Hypersim. We assume this is because kitchen contains many cluttered small objects, such as bottles, kitchenware, and utensils in the near field. SimSIN uses Habitat simulator \cite{savva2019habitat}, which renders images from synthetic (Replica \cite{straub2019replica}) or scanned but incomplete mesh (Matterport3D \cite{chang2017matterport3d} and HM3D \cite{ramakrishnan2021habitat}). Hypersim is pure synthetically rendered from delicately modeled spaces. Those simulation strategies cannot faithfully reflect high complexity of cluttered and small objects in real scenes. 
Therefore, they attain lower $\delta_1$, which indicates how object shapes are correctly estimated in the depth domain.
This serves as an understanding of simulation versus real data, showing a gap still exists to transfer knowledge well to real scenes.

The above studies consider InSpaceType as a testing set.
To further validate InSpaceType, we further create a training set for InSpaceType. The training set includes all 40K images except 1260 evaluation images and their nearby 2 frames.  We experiment with training on InSpaceType, NYUv2 \cite{silberman2012indoor}, and Hypersim \cite{roberts2021hypersim} using DPT-Hybrid (initialized from pretrained weights) and test on Replica \cite{chabra2019stereodrnet} and VA \cite{wu2022toward}. Results in Table \ref{table:add_1} show better zero-shot cross-dataset generalization for the introduced InSpaceType. 

\begin{table}[bt!]
\centering
\caption{\textbf{Zero-Shot generalization.} Hypersim, NYUv2, and InSpaceType are adopted as training sets. Replica (Left Table) and VA (Right Table) are used as indoor testing sets. Training on InSpaceType induces better results to validate InSpaceType's quality for zero-shot cross-dataset scenarios.}
\vspace{-6pt}
\tiny
\begin{tabular}{|c|c|c|c|c|c|c|c|c|c|c|c|}
\hline
\cellcolor[HTML]{99CCFF} Test on Replica & \cellcolor[HTML]{FAE5D3} AbsRel &  \cellcolor[HTML]{FAE5D3} RMSE & \cellcolor[HTML]{D5F5E3} $\delta_1$ & \cellcolor[HTML]{D5F5E3} $\delta_2$ & \cellcolor[HTML]{D5F5E3} $\delta_3$ & \cellcolor[HTML]{99CCFF} Test on VA & \cellcolor[HTML]{FAE5D3} AbsRel &  \cellcolor[HTML]{FAE5D3} RMSE & \cellcolor[HTML]{D5F5E3} $\delta_1$ & \cellcolor[HTML]{D5F5E3} $\delta_2$ & \cellcolor[HTML]{D5F5E3} $\delta_3$ \\
\hline
Hypersim & 0.1547 &  0.3833 & 79.88 & 92.94 & 97.39 & Hypersim &  0.1620 & 0.2997 & 77.65 & 94.53 & 98.29\\
%HM3D & 0.1517 & 0.3634 & 80.21 & 92.99 & 96.89 & HM3D & 0.2243 & 0.7016 & 68.46 & 89.54 & 96.39 \\
NYUv2 &  0.1524 & 0.3652 & 80.62 & 93.11 & 97.65 & NYUv2  & 0.1584 & 0.2650 & 80.19 & 95.21 & 98.78 \\
InSpaceType  & \textbf{0.1441} & \textbf{0.3347} & \textbf{81.82} & \textbf{93.51} & \textbf{98.12} & InSpaceType  & \textbf{0.1507} & \textbf{0.2483} & \textbf{81.74} & \textbf{95.50} & \textbf{99.01} \\
\hline
\end{tabular}
\label{table:add_1}
\vspace{-5pt}
\end{table}
\section{Intra-Dataset Study}
\label{gen_type}

In Section \ref{benchmark}, we emphasize cross-dataset benchmarks with space types. We benchmark several high-performing methods with a breakdown into space types. Further, we unveil and enumerate potential bias in several training sets. Next, we focus on InSpaceType itself for deeper analysis.

\textbf{[\RNum{4}: Dataset Fitting]} We use InSpaceType training set and train on small-size ConvNeXt network using standard $L_2$ loss supervised by groundtruth depth and test on the evaluation set. We use ConvNeXt networks for their high performance and as general-purpose networks to investigate data fitting and bias mitigation without loss of generality. 
%To compensate for different type frequency in training data, we implement a balanced type sampler to sample from all types with the same occurrences. 
Results in Table ~\ref{table:6} show most space types can fit in well when training on all types together. 
Large room and lounge are large-size spaces and naturally result in slightly higher RMSE. Kitchen's $\delta_1$ is a bit lower than other types due to the reason specified in [\RNum{3}]. It is worth noting that there is an apparent trend: for errors, larger rooms and longer ranges tend to have a higher estimation error; for accuracy, arbitrarily arranged small objects in the near field are challenging, a frequent scenario for kitchen. 

\begin{table}[bt!]
\centering
\caption{\textbf{Results of training and evaluation on train/eval splits of InSpaceType.} }
\vspace{-5pt}
\ssmall
\label{table:6}
\begin{tabular}{|c|c|c|c|c|c|c|}
\hline
 & Private room & Office & Hallway & Lounge & Meeting room & Large room \\
\hline
\cellcolor[HTML]{FAE5D3} RMSE & 0.1344 & 0.1729 & 0.2354 & 0.3185 & 0.1778 & 0.3153 \\
\cellcolor[HTML]{D5F5E3} $\delta_1$ & 98.41 & 96.93 & 95.81 & 96.88 & 98.11 & 98.22 \\
\hline
 & Classroom & Library & Kitchen & Playroom & Living room & Bathroom \\
\hline
\cellcolor[HTML]{FAE5D3} RMSE & 0.1725 & 0.2543 & 0.1825 & 0.1707 & 0.1556 & 0.0943 \\
\cellcolor[HTML]{D5F5E3} $\delta_1$ & 98.77 & 97.34 & 94.63 & 96.86 & 98.14 & 96.57 \\
\hline
\end{tabular}
\vspace{-9pt}
\end{table}

\textbf{[\RNum{5}: Mitigation of Uneven Distribution]} We also experiment with several basic and popular strategies to help mitigate the imbalance issue across different types. Specifically, we examine class re-weighting (CR) ~\cite{cui2019class,huang2016learning}, class-balanced sampling (CBS) \cite{kangdecoupling}, and Reptile-like meta-learning (ML) \cite{wu2023meta}. CR use weights inversely proportional to occurrences to compensate for types of higher occurrences. CBS is to sample from all classes with equal probability. Reptile ML uses bi-level optimization for learning to learn across tasks to attain higher generalizability and may mitigate unbalance.
We adopt ConvNeXt-small (Conv-sml) and ConvNeXt-base (Conv-b) as backbones and experiment on InSpaceType train and evaluation set. From Table ~\ref{table:7}, one can find CBS and ML are better strategies to attain lower standard deviation across types (t-STD) and better overall performance. Though CR attains lower t-STD, its overall performance drop as well. This is because CR could harm head-class performances as observed in literature ~\cite{wei2022open, zhou2020bbn}.

\begin{table}[bt!]
\centering
\caption{\textbf{Comparison for imbalance mitigation strategies.} Class weights (CR), class-balanced sampling (CBS), and meta-learning (ML) are examined. Type standard deviation (t-STD) of RMSE and $\delta_1$ computes standard deviation on twelve types. Higher t-STD indicates higher performance variation or more imbalance across types. }
\vspace{-6pt}
\footnotesize
\label{table:7}
\begin{tabular}{|c|c|c|c|c|c|c|c|c|c|}
\hline
\cellcolor[HTML]{99CCFF} Model & \cellcolor[HTML]{99CCFF}  Strategy & \cellcolor[HTML]{FAE5D3} AbsRel & \cellcolor[HTML]{FAE5D3} SqRel & \cellcolor[HTML]{FAE5D3} RMSE & \cellcolor[HTML]{D5F5E3} $\delta_1$ & \cellcolor[HTML]{D5F5E3} $\delta_2$ & \cellcolor[HTML]{D5F5E3} $\delta_3$ & \cellcolor[HTML]{EBDEF0} t-STD$_{\text{RMSE}}$ & \cellcolor[HTML]{EBDEF0} t-STD$_{\delta_1}$ \\
\hline
Conv-sml & w/o  & 0.0542 & 0.0181 & 0.1918  & 97.66 & 99.60 & 99.88  & 0.0642 & 1.4850 \\
Conv-sml & w/ CR & 0.0606 & 0.0202 & 0.2071 & 97.06 & 99.52 & 99.86 &  0.0630 & 1.4371\\ 
Conv-sml & w/ CBS & 0.0501 & 0.0166 & 0.1816 & 98.04 & 99.66 & 99.90 & 0.0600 & \textbf{1.1632}\\
Conv-sml & w/ ML & 0.0482 & 0.0160 & 0.1769 & 98.21 & 99.68 & 99.89 & \textbf{0.0580} & 1.3829\\
\hline
\hline
Conv-b &  w/o & 0.0510 & 0.0174 & 0.1846 & 97.96 & 99.63 & 99.89 & 0.0673 & 1.2236 \\
Conv-b & w/ CR & 0.0567 & 0.0196 & 0.1986 & 97.47 & 99.58 & 99.87  & 0.0619 & 1.1577 \\
Conv-b & w/ CBS & 0.0439 & 0.0146 & 0.1667 & 98.52 & 99.73 & 99.90 & 0.0561 & \textbf{1.0990}\\
Conv-b & w/ ML & 0.0451 & 0.0156 & 0.1692 & 98.44 & 99.70 & 99.90 & \textbf{0.0540} & 1.1269\\
\hline

\end{tabular}
\vspace{-8pt}
\end{table}

\textbf{[\RNum{6}: Generalization to Unseen Types]} In addition to generalization to unseen datasets, we are also curious about generalization to unseen types. We next divide the whole InSpaceType training set into different splits, train on each division, and then evaluate on InSpaceType eval split. The whole training set is divided into three groups (G) based on types. G1: private room, kitchen, living room, bathroom; G2: office, hallway, meeting room, classroom; G3: lounge, large room, playroom, library.
G1 is for household spaces; G2 is related to work or studies; G3 contains longer-range spaces. 
Models have only seen data from their respective group at training. For instance, a model trained on G1 cannot access RGBD pairs of classroom or hallway during training.
%These experiments hint at how training on some space types generalizes to other unseen types.
The categorization is based on similarity between types and concerns a situation where one collects training data almost in the same functionality that matches the primary application scenarios without considering different user scenarios. This is frequently encountered in scope-focused applications. Such as gaming VR systems are primarily for households, but the performance may drop with outlier use cases in classroom or workplace.
Results in Table ~\ref{table:15} left half show generalization to other types, and the right half shows evaluation on different depth ranges. 
Three depth ranges are defined: close, medium, and far. Close: a scene whose maximal depth is approximately within 5 meters. Medium: a scene whose maximal depth is approximately within 5-10 meters. Far: a scene whose maximal depth is approximately within 10-20 meters. The average maximal depth value is 3.78m for G1, 5.49m for G2, and 12.08m for G3.
We present another training set categorization based on ranges in Supplementary.

\begin{table}[bt!]
\centering
\caption{\textbf{Performance for training and evaluating on different groups.} G1$\to$ specifies the training group (G), and the entries below are evaluation groups. Three depth ranges: close, medium, and far are used to evaluate performances on scenes of different scales. See the text for the definition.}
\vspace{-6pt}
\tiny
\begin{tabular}{|c|c|c|c|c|c|c|c|c|c|c|c|c|c|}
\hline
\cellcolor[HTML]{99CCFF} G1$\to$Type & \cellcolor[HTML]{FAE5D3} AbsRel & \cellcolor[HTML]{FAE5D3} SqRel & \cellcolor[HTML]{FAE5D3} RMSE & \cellcolor[HTML]{D5F5E3} $\delta_1$ & \cellcolor[HTML]{D5F5E3} $\delta_2$ & \cellcolor[HTML]{D5F5E3} $\delta_3$ & \cellcolor[HTML]{99CCFF} G1$\to$Range & \cellcolor[HTML]{FAE5D3} AbsRel & \cellcolor[HTML]{FAE5D3} SqRel & \cellcolor[HTML]{FAE5D3} RMSE & \cellcolor[HTML]{D5F5E3} $\delta_1$ & \cellcolor[HTML]{D5F5E3} $\delta_2$ & \cellcolor[HTML]{D5F5E3} $\delta_3$ \\
\hline
G1 & 0.0511 & 0.0134 & 0.1461 & 98.12 & 99.71 & 99.92 & Close & 0.0984 & 0.0476 & 0.2877 & 89.48 & 98.08 & 99.66\\
G2 & 0.1607 & 0.1092 & 0.5501 & 77.22 & 94.91 & 98.90 & Medium & 0.1376 & 0.1241 & 0.5935 & 80.47 & 94.59 & 98.37\\
G3 & 0.2669 & 0.3851 & 1.1987 & 59.02 & 85.02 & 94.12 & Far & 0.2897 & 0.4375 & 1.3003 & 55.57 & 82.74 & 93.16 \\
%All & 0.1419 & 0.1355 & 0.5408 & 81.29 & 94.51 & 98.19 \\ 
\hline
\end{tabular}

\begin{tabular}{|c|c|c|c|c|c|c|c|c|c|c|c|c|c|}
\hline
\cellcolor[HTML]{99CCFF} G2$\to$Type & \cellcolor[HTML]{FAE5D3} AbsRel & \cellcolor[HTML]{FAE5D3} SqRel & \cellcolor[HTML]{FAE5D3} RMSE & \cellcolor[HTML]{D5F5E3} $\delta_1$ & \cellcolor[HTML]{D5F5E3} $\delta_2$ & \cellcolor[HTML]{D5F5E3} $\delta_3$ & \cellcolor[HTML]{99CCFF} G2$\to$Range & \cellcolor[HTML]{FAE5D3} AbsRel & \cellcolor[HTML]{FAE5D3} SqRel & \cellcolor[HTML]{FAE5D3} RMSE & \cellcolor[HTML]{D5F5E3} $\delta_1$ & \cellcolor[HTML]{D5F5E3} $\delta_2$ & \cellcolor[HTML]{D5F5E3} $\delta_3$ \\
\hline
G1 & 0.1418 & 0.0666 & 0.3497 & 82.29 & 96.81 & 99.39 & Close & 0.1063 & 0.0464 & 0.2720 & 87.97 & 97.80 & 99.59\\
G2 & 0.0673 & 0.0244 & 0.2250 & 96.33 & 99.45 & 99.86 & Medium & 0.1169 & 0.0728 & 0.4263 & 87.27 & 97.75 & 99.34\\
G3 & 0.2139 & 0.2485 & 0.9424 & 68.38 & 90.35 & 96.61 & Far & 0.2336 & 0.2886 & 1.0028 & 66.01 & 88.76 & 95.90\\
%All & 0.1311 & 0.0935 & 0.4420 & 84.22 & 96.28 & 98.92 \\ 
\hline
\end{tabular}

\begin{tabular}{|c|c|c|c|c|c|c|c|c|c|c|c|c|c|}
\hline
\cellcolor[HTML]{99CCFF} G3$\to$Type & \cellcolor[HTML]{FAE5D3} AbsRel & \cellcolor[HTML]{FAE5D3} SqRel & \cellcolor[HTML]{FAE5D3} RMSE & \cellcolor[HTML]{D5F5E3} $\delta_1$ & \cellcolor[HTML]{D5F5E3} $\delta_2$ & \cellcolor[HTML]{D5F5E3} $\delta_3$ & \cellcolor[HTML]{99CCFF} G3$\to$Range & \cellcolor[HTML]{FAE5D3} AbsRel & \cellcolor[HTML]{FAE5D3} SqRel & \cellcolor[HTML]{FAE5D3} RMSE & \cellcolor[HTML]{D5F5E3} $\delta_1$ & \cellcolor[HTML]{D5F5E3} $\delta_2$ & \cellcolor[HTML]{D5F5E3} $\delta_3$ \\
\hline
G1 & 0.1902 & 0.1416 & 0.4776 & 71.95 & 93.10 & 98.09 & Close & 0.2184 & 0.2070 & 0.5838 & 66.78 & 90.83 & 97.38\\
G2 & 0.1967 & 0.1657 & 0.6116 & 69.16 & 92.38 & 98.04 & Medium & 0.2129 & 0.2243 & 0.7016 & 68.46 & 89.54 & 96.39 \\
G3 & 0.0727 & 0.0395 & 0.3473 & 95.79 & 99.14 & 99.71 & Far & 0.0819 & 0.0491 & 0.3784 & 94.08 & 98.65 & 99.50 \\
%All & 0.1652 & 0.1267 & 0.4966 & 76.48 & 94.24 & 98.45 \\ 
\hline
\end{tabular}
\label{table:15}
\vspace{-15pt}
\end{table}

\begin{figure*}[bt!]
    \centering
    \includegraphics[width=0.94\linewidth]{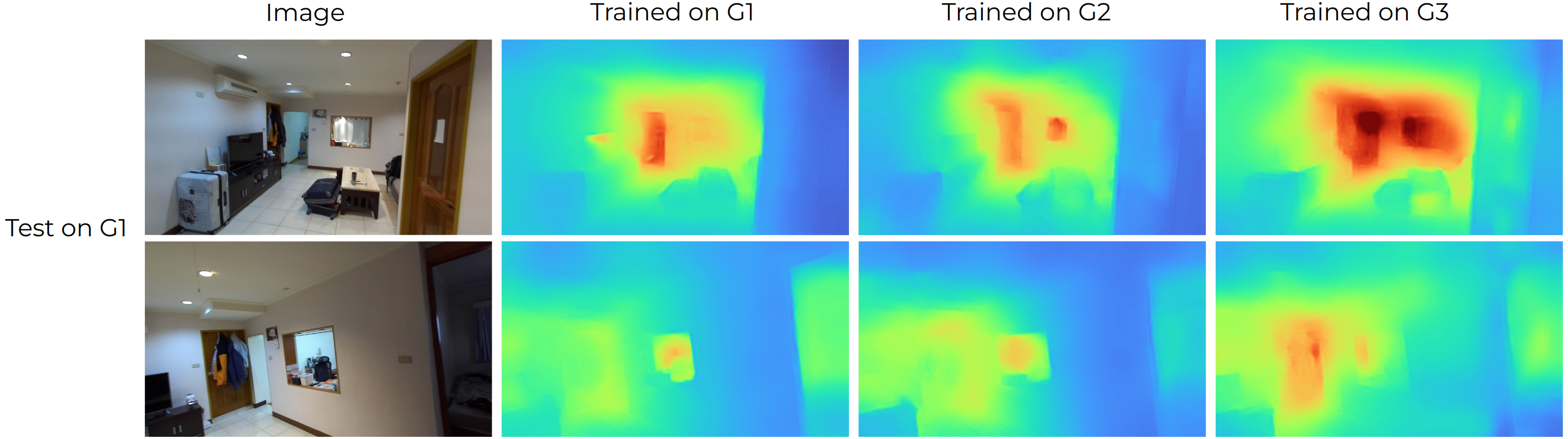}
    \vspace{-9pt}
    \includegraphics[width=0.94\linewidth]{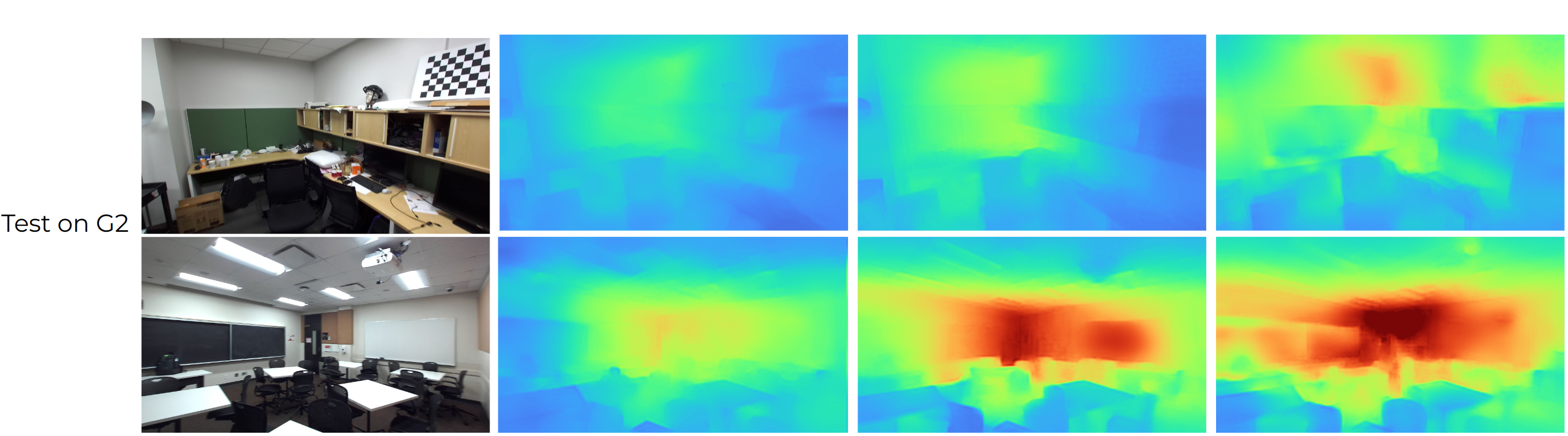}
    \vspace{-9pt}
    \includegraphics[width=0.94\linewidth]{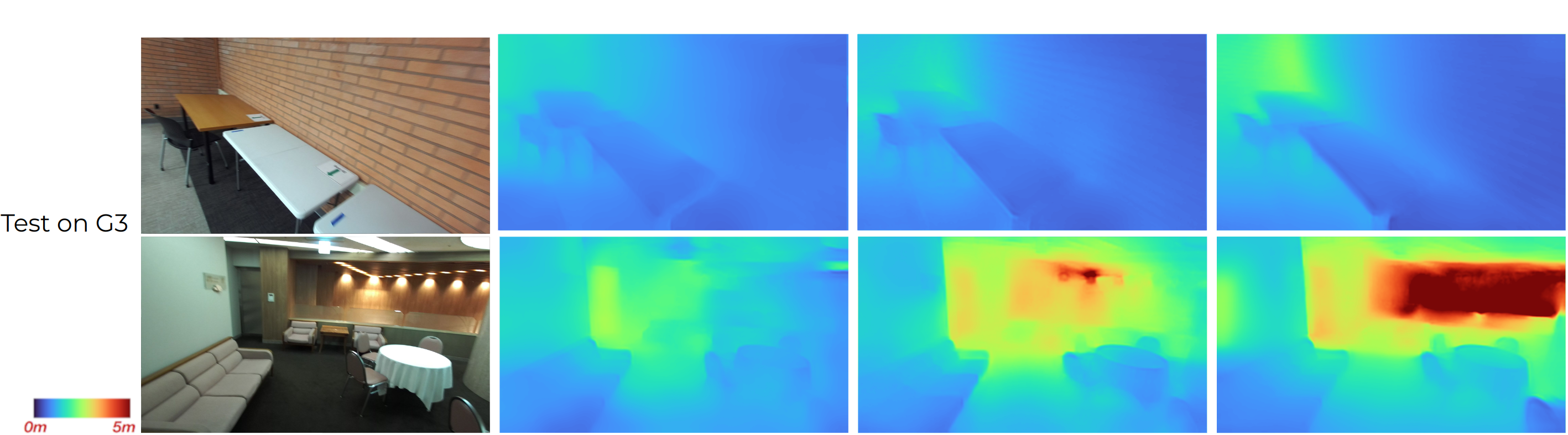}
    \vspace{-0pt}
    \caption{\textbf{Visualization of cross-group generalization.} }
    \vspace{-10pt}
    \label{fig_crossgp}
\end{figure*}

Training on specific groups can produce good performance on its dedicated types. However, one can observe training on only some types encounters severe issues in generalization to other unseen types, which further exhibit high variation between different indoor environments, and pretrained knowledge on some types may not easily transfer to other types.
For example, training on G1's household spaces cannot generalize to large or spacious room (G1$\to$G3 or Far) and show higher RMSE and lower $\delta_1$. Most indoor training datasets, such as NYUv2 or simulation from Matterport3D or Replica, are mostly curated for household spaces or smaller rooms. This may serve the needs of applications mainly for private room, but it potentially poses a training set bias towards close-range estimation, and the models trained on these datasets cannot be deployed to address different scenarios. Besides, we also observe training on large or spacious spaces (G3) can attain a bit better generalization to smaller rooms than the reverse setting, comparing between G3$\to$G1 and G1$\to$G3 or between G3$\to$Close and G1$\to$Far. We visualize the cross-group generalization result in Fig.~\ref{fig_crossgp}.

\section{Conclusion}
\label{conclusion}
\vspace{-7pt}
Unlike previous methods that focus on algorithmic developments, we are the first work to consider space types in indoor monocular depth estimation for robustness and practicability in deployment. We point out limitations in previous evaluations where performance variances across types are overlooked and present a novel dataset, InSpaceType, along with a hierarchical space type definition to facilitate our study. We give thorough studies to analyze and benchmark performance based on space types. 
Ten high-performing methods are examined, and we find they suffer from severe performance imbalance between space types. 
We analyze a total of 4 training datasets and enumerate their strength and weakness space types. 3 popular strategies, namely, class reweighting, type-balanced sampler, and meta-learning, are studied to mitigate imbalance.
Further, we find generalization to unseen space types challenging due to high diversity of objects and mismatched scales across types. Overall, this work pursues a practical purpose and emphasizes the importance of this usually overlooked factor- space type in indoor environments. We call for attention to safety concerns for model deployment without considering performance variance across space types.

\textbf{Limitations}. This work only considers monocular depth estimation. Other popular scopes for depth estimation such as outdoor domain, stereo approaches, or multiview scene reconstruction may also suffer from performance imbalance across different types. We choose to operate on monocular depth estimation since it is the most fundamental task with only an image needed and is especially widely useful in many recent popular applications or deployed systems such as indoor AR on smartphones, VR gaming, novel view synthesis, or video generation for indoor scenes. 
This work specifically zooms into the factor space type, but this may not be the only important factor blocking the generalization.

{\small
\bibliographystyle{ieee_fullname}
\bibliography{egbib}
}

\end{document}